\documentclass[runningheads]{llncs}
\usepackage{graphicx}
\usepackage{amsmath,amssymb} 
\usepackage{color}
\usepackage{amssymb}
\usepackage{url}            
\usepackage{multirow}

\newcommand{\zy}[1]{{\color{black}{#1}}}

\let\on=\operatorname
\newcommand{\ud}{\,\mathrm{d}}

\begin{document}
\title{Anatomical Data Augmentation \\ via Fluid-based Image Registration} 
%
%
\author{}
\author{Zhengyang Shen \and
	Zhenlin Xu \and
	Sahin Olut \and
	Marc Niethammer}

\institute{Department of Computer Science, UNC Chapel Hill}
\authorrunning{Z. Shen et al.}

\maketitle              
\begin{abstract}
We introduce a fluid-based image augmentation method for medical image analysis. In contrast to existing methods, our framework generates anatomically meaningful images via interpolation from the geodesic subspace underlying given samples. Our approach consists of three steps: 1) given a source image and a set of target images, we construct a geodesic subspace using the Large Deformation Diffeomorphic Metric Mapping (LDDMM) model; 2) we sample transformations from the resulting geodesic subspace; 3) we obtain deformed images and segmentations via interpolation. Experiments on brain (LPBA) and knee (OAI) data illustrate the performance of our approach on two tasks: 1) data augmentation during training and testing for image segmentation; 2) one-shot learning for single atlas image segmentation. We demonstrate that our approach generates anatomically meaningful data and improves performance on these tasks over competing approaches. Code is available at \url{https://github.com/uncbiag/easyreg}.

\end{abstract}
  \section{Introduction}
Training data-hungry deep neural networks is challenging for medical image analysis where manual annotations are more difficult and expensive to obtain than for natural images. Thus it is critical to study how to use scarce annotated data efficiently, e.g., via data-efficient models~\cite{vakalopoulou2018atlasnet,heinrich2018obelisk}, training strategies~\cite{paschali2019data} and semi-supervised learning strategies utilizing widely available unlabeled data through self-training~\cite{bai2017semi,nie2018asdnet},  regularization~\cite{baur2017semi}, and multi-task learning~\cite{chen2019multi,xu2019deepatlas,zhou2019collaborative}.

An alternative approach is data augmentation. Typical methods for medical image augmentation include random cropping~\cite{hussain2017differential}, geometric transformations~\cite{oliveira2017augmenting,milletari2016v,roth2015anatomy} (e.g., rotations, translations, and free-form deformations), and photometric (i.e., color) transformations~\cite{learned2005data,pereira2016brain}. Data-driven data augmentation has also been proposed, to \zy{learn generative models for synthesizing images with new appearance~\cite{shin2018medical,frid2018gan}, to estimate class/template-dependent distributions of deformations~\cite{hauberg2016dreaming,park2018representing,zhang2013bayesian} or both~\cite{zhao2019data,chaitanya2019semi}.} Compared with these methods, our approach focuses on a geometric view and constructs a continuous geodesic subspace as an estimate of the space of anatomical variability.

\begin{figure}[!t]
\centering
\includegraphics[width=0.9\textwidth]{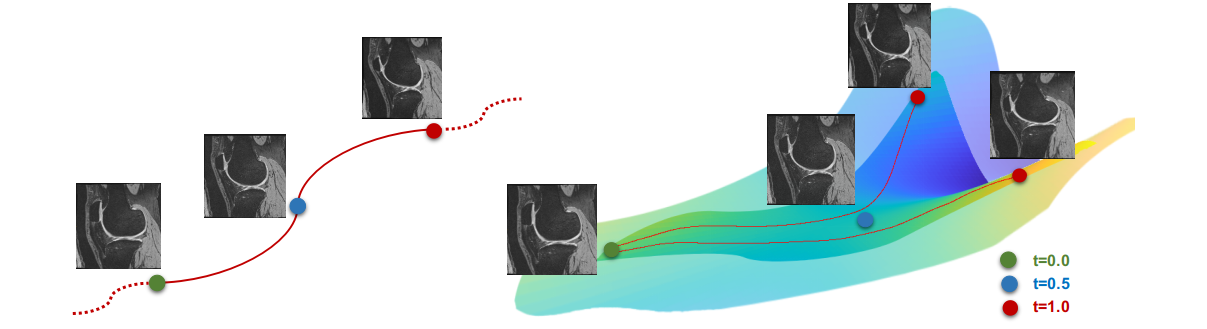}
\caption{Illustration of our fluid-based data augmentation using a 1D (left) and 2D (right) geodesic subspace. We assume a registration from a source to a target image in unit time. In 1D, we can sample along the geodesic path ($t\in[0,1]$) between the source ($t=0$) and the target images ($t=1$). We can also extrapolate $t\notin[0,1]$.  In the 2D case, a source and two target images define a two-dimensional geodesic subspace.} 
\label{fig:1d_2d_space}
\end{figure}

Compared with the high dimensionality of medical images, anatomical variability is often assumed to lie in a much lower dimensional space~\cite{aljabar2012manifold}. Though how to directly specify this space is not obvious, we can rely on reasonable assumptions informed by the data itself. We assume there is a diffeomorphic transformation between two images, that image pairs can be connected via a geodesic path, and that appearance variation is implicitly captured by the appearance differences of a given image population. For longitudinal image data, we can approximate images at   intermediate time points by interpolation or predict via extrapolation. As long as no major appearance changes exist, diffeomorphic transformations can provide realistic intermediate images\footnote{In some cases, for example for lung images, sliding effects need to be considered, violating the diffeomorphic assumption.}. Based on these considerations, we propose a data augmentation method based on fluid registration which produces {\it anatomically plausible} deformations and retains appearance differences of a given image population. 
Specifically, we choose the Large Deformation Diffeomorphic Metric Mapping (LDDMM) model as our fluid registration approach. LDDMM comes equipped with a metric and results in a geodesic path between a source and a target image which is parameterized by the LDDMM initial momentum vector field. 
Given two initial momenta in the tangent space of the same source image, we can define a geodesic plane, illustrated in Fig.~\ref{fig:1d_2d_space}; similarly, we can construct higher dimensional subspaces based on convex combinations of sets of momenta~\cite{qiu2011principal}. Our method includes the following steps: 1) we compute a set of initial momenta for a source image and a set of target images; 2) we generate an initial momentum via a convex combination of initial momenta; 3) we sample a transformation on the geodesic path determined by the momentum; and 4) we warp the image and its segmentation according to this transformation.

\zy{Data augmentation is often designed for the training phase. However, we show the proposed approach can be extended to the testing phase, e.g., a testing image is registered to a set of training images (with segmentations) and the deep learning (DL) segmentation model is evaluated in this warped space (where it was trained, hence ensuring consistency of the DL input); the predicted segmentations are then mapped back to their original spaces. In such a setting, using LDDMM can guarantee the existence of the inverse map whereas traditional elastic approaches cannot.
}

\noindent
{\bf Contributions:} 1) We propose a general fluid-based approach for anatomically consistent medical image augmentation \zy{for both training and {\it testing}}. 2) We build on LDDMM and can therefore assure well-behaved diffeomorphic transformations when interpolating {\it and} extrapolating samples with large deformations. 3) Our method easily integrates into different tasks, such as segmentation and one-shot learning for which we show general performance improvements.

\section{LDDMM Method}
\label{sec:lddmm}

LDDMM~\cite{beg2005computing} is a fluid-based image registration model, estimating a spatio-temporal velocity field $v(t,x)$ from which the spatial transformation $\varphi$ can be computed via integration of $\partial_t \varphi (t,x) = v(t,\varphi (t,x)),~\varphi(0,x)=x\,.$ For appropriately regularized velocity fields~\cite{dupuis1998variational}, diffeomorphic transformations can be guaranteed. The optimization problem underlying LDDMM can be written as 
\begin{equation}
v^* = \underset{v}{\text{argmin}}~ \frac 12 \int_0^1 \|v(t) \|^2_L \ud dt + \on{Sim}(I(1),I_1) \quad\text{s.t.}\quad
\partial_t I + \langle \nabla I, v\rangle=0,~ I(0)=I_0\enspace,
\label{eq:lddmm}
\end{equation}
where $\nabla$ denotes the gradient, $\langle \cdot,\cdot\rangle$ the inner product, and $\on{Sim}(A,B)$ is a similarity measure between images.
We note that $I(1,x)=I_0\circ\varphi^{-1}(1,x)$, where $\varphi^{-1}$ denotes the inverse of $\varphi$ in the target image space. The evolution of this map follows $\partial_t \varphi^{-1} + D\varphi^{-1}v=0$, 
where $D$ is the Jacobian. Typically, one seeks a velocity field which deforms the source to the target image in unit time. To assure smooth transformations, LDDMM penalizes non-smooth velocity fields via the norm $\|v\|_L^2=\langle Lv,Lv\rangle$, where $L$ is a differential operator.

At optimality the following equations hold~\cite{younes2009evolutions} and the entire evolution can be parameterized via the initial vector-valued momentum,  $m=L^\dagger L v$: 
\begin{eqnarray}
&~&m(0)^* = \underset{m(0)}{\text{argmin}}~ \frac 12 \langle m(0),v(0)\rangle  + \operatorname{Sim}(I_0\circ\varphi^{-1}(1),I_1),\label{eq:lddmm_shooting_energy}\\ 
&~&\quad\text{s.t.}\quad\varphi_t^{-1} + D\varphi^{-1}v=0, 
\quad \varphi(0,x)=x\enspace,\label{eq:lddmm_shooting_map_advection}\\
&~&\partial_t m + \on{div}(v) m + D v^T(m) + D m(v) = 0,~m(0)=m_0, v=K\star m\,,
\label{eq:lddmm_epdiff}
\end{eqnarray}
where we assume $(L^\dagger L)^{-1}m$ is specified via convolution $K\star m$. Eq.~\ref{eq:lddmm_epdiff} is the Euler-Poincar\'e equation for diffeomorphisms (EPDiff)~\cite{younes2009evolutions},  defining the evolution of the spatio-temporal velocity field based on the initial momentum $m_0$. 

The geodesic which connects the image pair $(I_0,I_0\circ\varphi^{-1}(1))$ and approximates the path between $(I_0,I_1)$ is specified by $m_0$. We can sample along the geodesic path, assuring diffeomorphic transformations.
As LDDMM assures diffeomorphic transformations, we can also obtain the inverse transformation map, $\varphi$ (defined in source image space, whereas $\varphi^{-1}$ is defined in target image space) by solving 
\begin{equation}
\varphi(1,x)= x + \int_0^1 v(t,\varphi(t,x))~dt,~\varphi(0,x)=x.
\label{eq:inverse_map}
\end{equation}
Computing the inverse for an arbitrary displacement field on the other hand requires the numerical minimization of $\|\varphi^{-1} \circ \varphi - id\|^2$. Existence of the inverse map cannot be guaranteed for such an arbitrary displacement field.

\section{Geodesic Subspaces}
\label{sec:geos}

We define a geodesic subspace constructed from a source image and a set of target images. Given a dataset of size N, $I_c\in\mathbb{R}^D$ denotes an individual image $c\in\{1\dots N\}$, where $D$ is the number of voxels. For each source image $I_c$, we further denote a target set of $K$ images as ${\bf I^c_K}$.
 We define $M_K^c:=\{m_0^{cj}|\mathcal{M}(I_c,I_j), \mathcal{M}:\mathbb R^D\times \mathbb R^D\xrightarrow{}\mathbb R^{D\times d},I_j\in{\bf I^c_K}\}$ as a set of K different initial momenta, where $\mathcal M$ maps from an image pair to the corresponding initial momentum via Eqs.~\ref{eq:lddmm_shooting_energy}-\ref{eq:lddmm_epdiff}; $d$ is the spatial dimension. We define convex combinations of $M_K^c$ as 
 \begin{equation}
 C(M_K^c) := \left\{\tilde m_0^c \bigg\rvert \tilde m_0^c =\sum_{j=1}^{K} \lambda_{j} m_0^{cj}, m_0^{cj}\in M_K^c, \lambda_{j} \geq 0~\forall j,  \sum_{j=1}^{K} \lambda_{j}=1\right\}.
 \label{eq:convex}
 \end{equation}
 Restricting ourselves to convex combinations, instead of using the entire space defined by arbitrary linear combinations of the momenta $M_k^c$ allows us to retain more control over the resulting momenta magnitudes. For our augmentation strategy we simply sample an initial momentum $\tilde{m}_0^c$ from $C(M_K^c)$, which, according to the EPDiff Eq.~\ref{eq:lddmm_epdiff}, determines a geodesic path starting from $I_c$. 
 If we set $K=2$, for example, the sampled momentum parameterizes a path from a source image toward two target images, where the $\lambda_i$ weigh how much the two different images drive the overall deformation. As LDDMM registers a source to a target image in unit time, we obtain interpolations by additionally sampling $t$ from $[0,1]$, resulting in the intermediate deformation $ \varphi_{\tilde m_0^c}^{-1}(t)$ from the geodesic path starting at $I_c$ and determined by ${\tilde m_0^c}$. We can also extrapolate by sampling $t$ from $\mathbb{R}\setminus[0,1]$. We then synthesize images via interpolation: $I_c \circ \varphi_{\tilde m_0^c}^{-1}(t)$.

\section {Segmentation}
\label{sec:segmentation}
In this section, we first introduce an augmentation strategy for general image segmentation (Sec.~\ref{sec:general_segmentation}) and then a variant for one-shot segmentation (Sec.~\ref{sec:one_shot_segmentation}).

\subsection{Data augmentation for general image segmentation}
\label{sec:general_segmentation}

We use a two-phase data augmentation approach consisting of (1) pre-augmentation of the training data and (2) post-augmentation of the testing data.
During the training phase, for each training image, $I_c$, we generate a set of new images by sampling from its geodesic subspace, $C(M_K^c)$. This results in a set of deformed images which are anatomically meaningful and retain the appearance of $I_c$. We apply the same sampled spatial transformations to the segmentation of the training image, resulting in a new set of warped images and segmentations. We train a segmentation network based on this augmented dataset.

During the testing phase, for each testing image, we also create a set of new images using the same strategy described above. Specifically, we pair a testing image with a set of {\it training} images to create the geodesic subspace for sampling. This will result in samples that come from a similar subspace that has been used for augmentation during training. A final segmentation is then obtained by warping the predicted segmentations back to the original space of the image to be segmented and applying a label-fusion strategy. Consequently, we expect that the segmentation network performance will be improved as it (1) is allowed to see multiple views of the same image and (2) the set of views is consistent with the set of views that the segmentation network was trained with.

\begin{figure}[!h]
\centering
\includegraphics[width=0.9\textwidth]{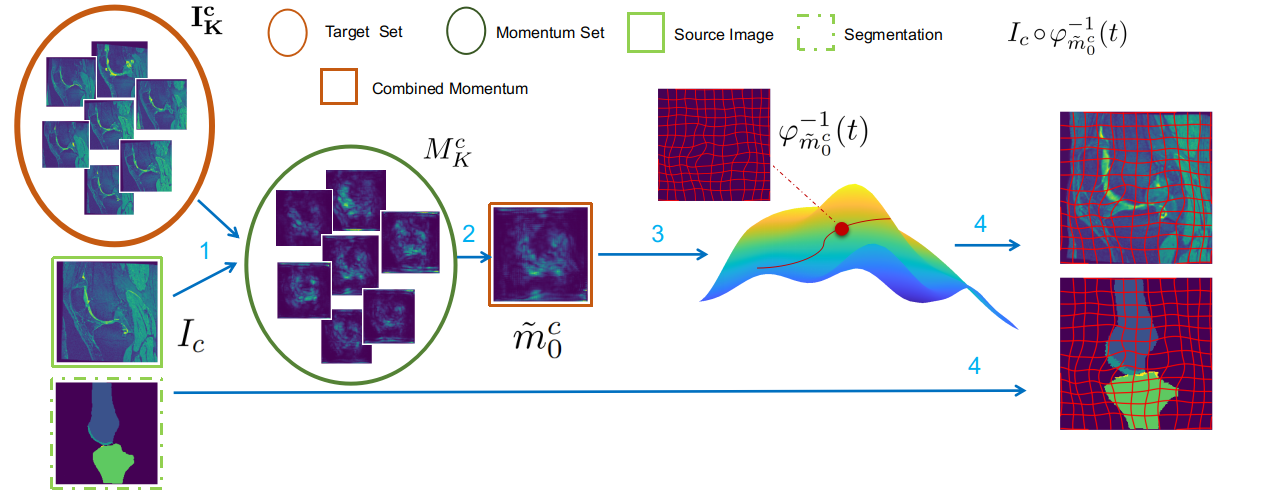}
\centering
\caption{Illustration of the training phase data augmentation. Given a source image $I_c$ and a set of target images ${\bf I^c_K}$, a set of momenta $M_K^c$ is first computed. Then a momentum $\tilde m_0^c$ is sampled from the convex combination of these momenta $C(M_K^c)$ defining a geodesic path starting from the source image. Lastly, a transformation $\varphi_{\tilde m_0^c}^{-1}(t)$ is sampled on the geodesic and used to warp the source image and its segmentation.}
\label{fig:train_phase}
\end{figure}

Fig.~\ref{fig:train_phase} illustrates the training phase data augmentation. We first compute $M_K^c$ by picking an image $I_c$, $c \in \{1\dots N\}$ from a training dataset of size $N$ and a target set ${\bf I^c_K}$ of cardinality $K$, also sampled from the training set. We then sample $\tilde m_0^c \in C(M_K^c)$ defining a geodesic path from which we sample a deformation $\varphi_{\tilde m_0^c}^{-1}(t)$ at time point $t$. 
We apply the same strategy multiple times and obtain a new deformation set for each $I_c$, $c \in \{1\dots N\}$.
The new image set $\{I_c \circ \varphi_{\tilde m_0^c}^{-1}(t)\}$ consisting of the chosen set of random transformations of $I_c$
and the corresponding segmentations can then be obtained by interpolation.
 
\begin{figure}[!h]
\centering
\includegraphics[width=0.9\textwidth]{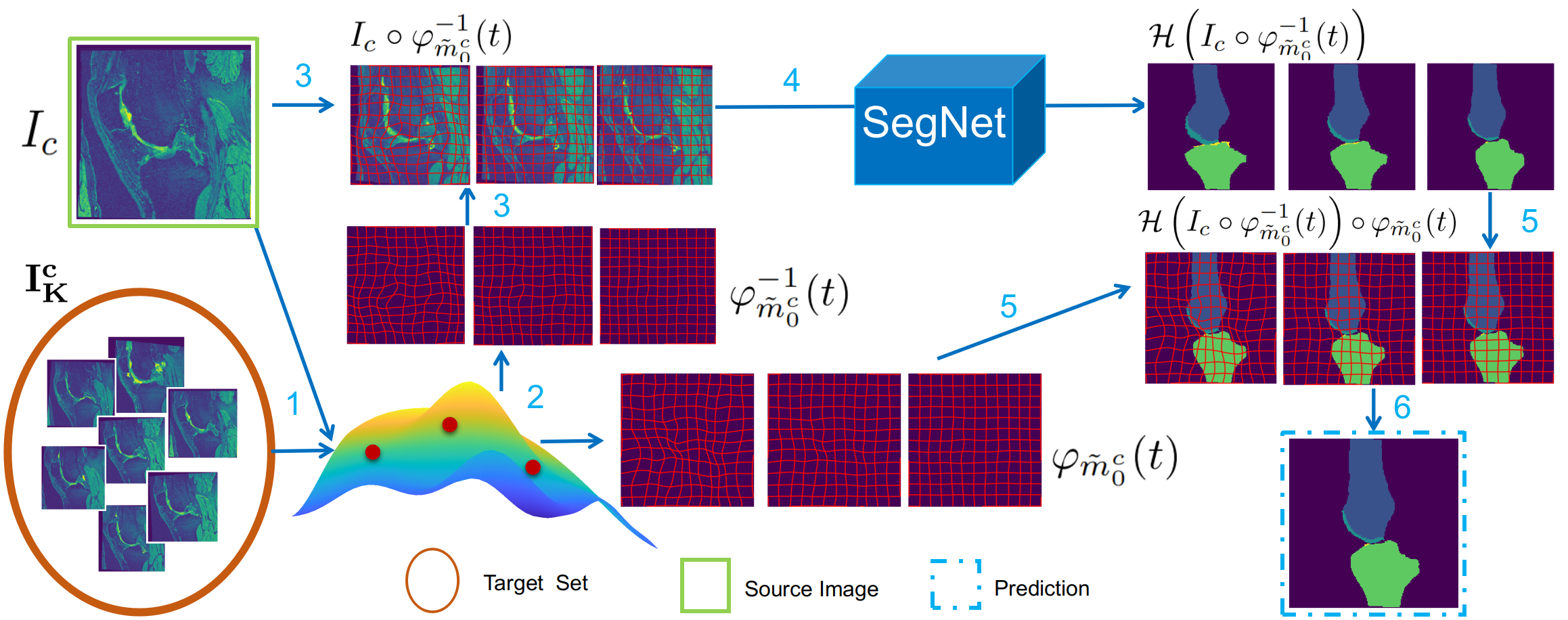}
\centering
\caption{Illustration of the testing phase data augmentation. Given a source image $I_c$ and a set of target images ${\bf I^c_K}$, a geodesic subspace is determined first. A set of transformations $\varphi_{\tilde m_0^c}^{-1}(t)$ is then sampled from this space and, at the same time, the corresponding inverse transformations $\varphi_{\tilde m_0^c}(t)$ are obtained. A segmentation network $\mathcal{H}$ is applied to each warped image and the resulting segmentations $\mathcal{H} (I_c \circ \varphi_{\tilde m_0^c}^{-1}(t))$ are warped back to the source image space. A label fusion strategy is applied to obtain the final segmentation. }
\label{fig:test_phase}
\end{figure}

Fig.~\ref{fig:test_phase} illustrates the testing phase data augmentation. For a test image $I_c$ and its target set ${\bf I^c_K}$ sampled from the {\it training set}, we obtain a set of transformations $\{\varphi_{\tilde m_0^c}^{-1}(t)\}$. 
By virtue of the LDDMM model these transformations are invertible. For each $\varphi_{\tilde m_0^c}^{-1}(t)$ we can therefore efficiently obtain the corresponding inverse map $\varphi_{\tilde m_0^c}(t)$. We denote our trained segmentation network by $\mathcal{H}:\mathbb{R}^D \xrightarrow{} \mathbb{R}^{D \times L}$ which takes an image as its input and predicts its segmentation labels. Here, $L$ is the number of segmentation labels. 
Each prediction $\mathcal{H}\left (I_c \circ \varphi_{\tilde m_0^c}^{-1}(t)\right)$ is warped back to the space of $I_c$ via $\mathcal{H}\left (I_c \circ \varphi_{\tilde m_0^c}^{-1}(t)\right )\circ\varphi_{\tilde m_0^c}(t)$. The final segmentation is obtained by merging all warped predictions via a label fusion strategy.

{\bf Dataset}
The LONI Probabilistic Brain Atlas~\cite{shattuck2008construction} (LPBA40) dataset contains volumes of 40 healthy patients with 56 manually annotated anatomical structures. We affinely register all images to a mean atlas~\cite{joshi2004unbiased}, resample to isotropic spacing of 1~mm, crop them to $196\times164\times196$ and intensity normalize them to $[0,1]$ via histogram equalization.
We randomly take 25 patients for training, 10 patients for testing, and 5 patients for validation.

The Osteoarthritis Initiative~\cite{oaidataset} (OAI) provides manually labeled knee images with segmentations of femur and tibia
as well as femoral and tibial cartilage~\cite{ambellan2019automated}. We first affinely register all images to a mean atlas~\cite{joshi2004unbiased}, resample them to isotropic spacing of 1~mm, and crop them to $160\times200\times200$.
We randomly take 60 patients for training, 25 patients for validation, and 52 patients for testing.

To evaluate the effect of data augmentation on training datasets with different sizes, we further sample 5, 10, 15, 20, 25 patients as the training set on LPBA40 and 10, 20, 30, 40, 60 patients as the training set for OAI.

{\bf Metric} We use the average Dice score over segmentation classes for all tasks in Sec.~\ref{sec:general_segmentation} and Sec.~\ref{sec:one_shot_segmentation}.

{\bf Baselines}
{\it Non-augmentation} is our lower bound method. We use a class-balanced random cropping schedule during training~\cite{xu2018contextual}.
We use this cropping schedule for all segmentation methods that we implement. We use a U-Net~\cite{ronneberger2015u} segmentation network. 
{\it Random B-Spline Transform} is a transformation locally parameterized by randomizing the location of B-spline control points. 
 Denote $(\cdot,\cdot)$ as the number of control points distributed over a uniform mesh and the standard deviation of the normal distribution, units are in $mm$. The three settings we use are $(10^3,3)$, $(10^3,4)$, $(20^3,2)$. During data augmentation, we randomly select one of the settings to generate a new example.

{\bf Settings} 
 During the training augmentation phase ({\it pre-aug}), we randomly pick a source image and $K$ targets, uniformly sample $\lambda_i$ in Eq.~\ref{eq:convex} and then uniformly sample $t$. For LPBA40, we set $K=2$ and $t\in[-1,2]$; for the OAI data, we set $K=1$ and $t\in[-1,2]$. For all training sets with different sizes, for both the B-Spline and the fluid-based augmentation methods and for both datasets, we augment the training data by 1,500 cases. During the testing augmentation phase ({\it post-aug}), for both datasets, we set $K=2$ and $t \in [-1,2]$ and draw 20 synthesized samples for each test image. The models trained via the augmented training set are used to predict the segmentations. To obtain the final segmentation, we sum the softmax outputs of all the segmentations warped to the original space and assign the label with the largest sum. We test using the models achieving the best performance on the validation set. We use the optimization approach in~\cite{niethammer2019_cvpr} and the network of~\cite{shen2019networks,shen2019region} to compute the mappings $\mathcal{M}$ on LPBA40 and OAI, respectively. 

\begin{figure}
  \begin{minipage}[b]{0.7\textwidth}
    \centering
\includegraphics[width=1.\textwidth]{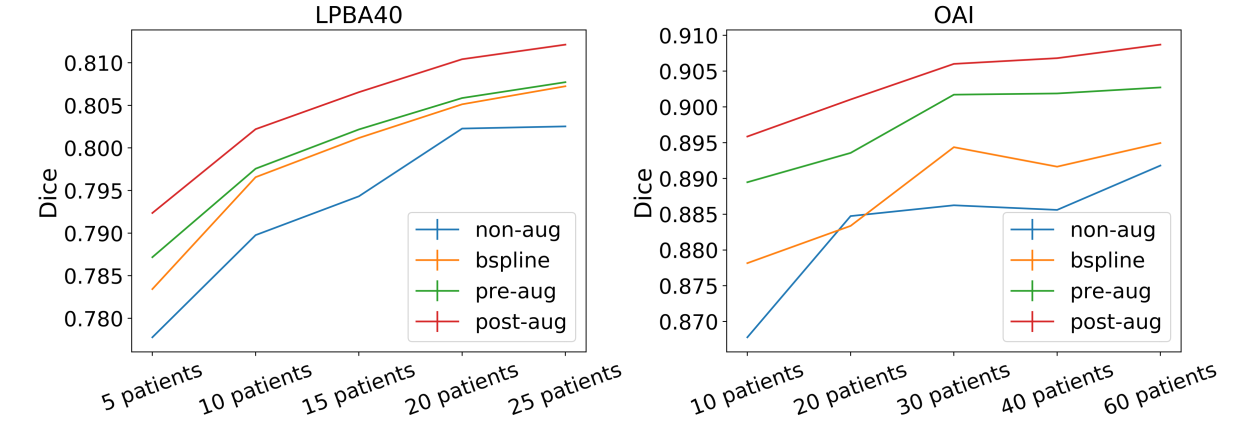}
    \par\vspace{-1.3cm}
  \end{minipage}%
  \begin{minipage}[b]{0.3\textwidth}
    \centering
\scalebox{0.8}{
\tabcolsep=0.6pt
\begin{tabular}{|cc|}
  \hline 
\multirow{2}{*}{} &
\multicolumn{1}{c|}{OAI Dataset}  \\
Method & Dice (std)\\\hline
$Brainstorm$ & 79.94 (2.22) \\
$Fluid\text{-}Aug$ &80.81 (2.35)\\
$Brainstorm_{real}$ &86.83 (2.21)  \\
$Fluid\text{-}Aug_{real_{t1}}$ &  87.74 (1.82)\\
$Fluid\text{-}Aug_{real}$ &  88.31 (1.56)\\
$Upper\text{-}bound$  & 90.01 (1.58)\\
\hline
\end{tabular}
}
\end{minipage}
\caption{\label{fig:seg_perf}  Segmentation performance for segmentation tasks. The left two plots show Dice scores for the different methods with different training set sizes on the LPBA40 and OAI datasets for general segmentation. Performance increases with training set size.  Fluid-based augmentation (pre-aug and post-aug)
shows the best performance. The right table compares the performance for one-shot segmentation in Sec.~\ref{sec:one_shot_segmentation}. Fluid-based augmentation methods perform better than their {\it Brainstorm} counterparts.
}
\end{figure}

{\bf Results}
Fig.~\ref{fig:seg_perf} shows the segmentation performance on the LPBA40 and the OAI datasets. For training phase augmentation, fluid-based augmentation improves accuracy over non-augmentation and B-Spline augmentation by a large margin on the OAI dataset and results in comparable performance on the LPBA40 dataset. This difference might be due to the larger anatomical differences in the \zy{LPBA40} dataset compared to the \zy{OAI} dataset; \zy{such large differences might not be well captured by inter- and extrapolation along a few geodesics.} Hence, the OAI dataset may benefit more from the anatomically plausible geodesic space. When test phase augmentation is used in addition to training augmentation, performance is further improved. This shows that the ensemble strategy used by post-aug, where the segmentation network makes a consensus decision based on different views of the image to be segmented, is effective. In practice, we observe that high-quality inverse transformations (that map the segmentations back to the test image space) are important to achieve good performance. These inverse transformations can efficiently be computed via Eq.~\ref{eq:inverse_map} for our fluid-based approach.

\subsection{Data augmentation for one-shot segmentation}
\label{sec:one_shot_segmentation}

 We explore one-shot learning. Specifically, we consider single atlas medical image segmentation, where only the atlas image has a segmentation, while all other images are unlabeled. We first review {\it Brainstorm}~\cite{zhao2019data}, a competing data augmentation framework for one-shot segmentation. We then discuss our modifications.
 
In {\it Brainstorm}, the appearance of a sampled unlabeled image is first transfered to atlas-space and subsequently spatially transformed by registering to another sampled unlabeled image. Specifically, a registration network $\mathcal H^r$ is trained to predict the displacement field
 between the atlas $A$ and the unlabeled images. For a given image $I_c$, the predicted transformation to $A$ is $\varphi_c(x)=\mathcal H^r(I_c, A)+x$.
 A set of approximated inverse transformations $\{\varphi_c^{-1},c\in1\dots N\}$ from the atlas to the image set can also be predicted by the network $\mathcal H^r$. These inverse transformations capture the anatomical diversity of the unlabeled set and are used to deform the images.
 Further, an appearance network $\mathcal H^a$ is trained to capture the appearance of the unlabeled set.
 The network is designed to output the residue $r$ between the warped image $I_c\circ\varphi$ and the atlas, $r_c =\mathcal H^a(A,I_c\circ \varphi_c)$. Finally, we obtain a new set of segmented images by applying the transformations to the atlas with the new appearance: $\{(A+r_i)\circ \varphi_j^{-1}, i,j \in 1\dots N\}$.  

 We modify the {\it Brainstorm} idea as follows: 1) Instead of sampling the transformation $\varphi_c^{-1}$, we sample $\varphi^{-1}_{\tilde m_0^c}(t)$ based on our fluid-based approach;
 2) We remove the appearance network and instead simply use $\{I_c \circ \varphi_c, c\in 1\dots N\}$ to model appearance. I.e., we retain the appearance of individual images, but deform them by going through atlas space. This results by construction in a realistic appearance distribution. Our synthesized images are $\{(I_i \circ \varphi_i)\circ \varphi^{-1}_{\tilde m_0^j}(t), \tilde{m}_0^j \in C(M_K^j),t\in\mathbb R, i,j \in 1\dots N\}$. We refer to this approach as $Fluid\text{-}Aug_{real}$.

{\bf Dataset} We use the OAI dataset with 100 manually annotated images and a segmented mean atlas~\cite{joshi2004unbiased}.
We only use the atlas segmentation for our one-shot segmentation experiments. 

{\bf Baseline}
{\it Upper-bound} is a model trained from 100 images and their manual segmentations. We use the same U-net as for the general segmentation task in Sec.~\ref{sec:general_segmentation}.
{\it Brainstorm} is our baseline. We train a registration network and an appearance network separately, using the same network structures as in~\cite{zhao2019data}. We sample a new training set of size 1,500 via random compositions of the appearance and the deformation.
We also compare with a variant replacing the appearance network, where the synthesized set can be written as $\{(I_i \circ \varphi_i)\circ \varphi_j^{-1}, i,j \in 1\dots N\}$. We refer to this approach as $Brainstorm_{real}$.

{\bf Settings}
We set $K=2$ and $t\in[-1,2]$ and draw a new training set with 1,500 pairs the same way as in Sec.~\ref{sec:general_segmentation}.
We also compare with a variant where we set $t=1$ (instead of randomly sampling it), which we denote $Fluid\text{-}Aug_{real_{t=1}}$. Further, we compare with a variant using the appearance network, where the synthesized set is $\{(A+r_i)\circ \varphi^{-1}_{\tilde m_0^j}(t), m_0^j \in C(M_K^j),t\in\mathbb R, i,j \in 1\dots N\}$. We refer to this approach as $Fluid\text{-}Aug$.

Fig.~\ref{fig:seg_perf} shows better performance for fluid-based augmentation than for {\it Brainstorm} when using either real or learnt appearance.
Furthermore, directly using the appearance of the unlabeled images shows better performance than using the appearance network. Lastly, randomizing over the location on the geodesic ($Fluid-Aug_{real}$) shows small improvements over fixing $t=1$ ($Fluid-Aug_{real_{t_1}}$).

\section{Conclusion}
\label{sec:concl}
We introduced a fluid-based method for medical image data augmentation. Our approach makes use of a geodesic subspace capturing anatomical variability. We explored its use for general segmentation and one-shot segmentation, achieving improvements over competing methods.
Future work will focus on efficiency improvements. Specifically, computing the geodesic subspaces is costly if they are not approximated by a registration network. We will therefore explore introducing multiple atlases to reduce the number of possible registration pairs.

~\\
\noindent
\textbf{Acknowledgements:} Research reported in this publication was supported by the National Institutes of Health (NIH) and the National Science Foundation (NSF) under award numbers NSF EECS1711776 and NIH 1R01AR072013. The content is solely the responsibility of the authors and does not necessarily represent the official views of the NIH or the NSF.

{\small
\bibliographystyle{abbrv}

}
\clearpage

\setcounter{page}{1}
\section{Supplementary Material}
\vspace{-1cm}
\begin{figure}[!htb]
\centering
\includegraphics[width=0.95\textwidth]{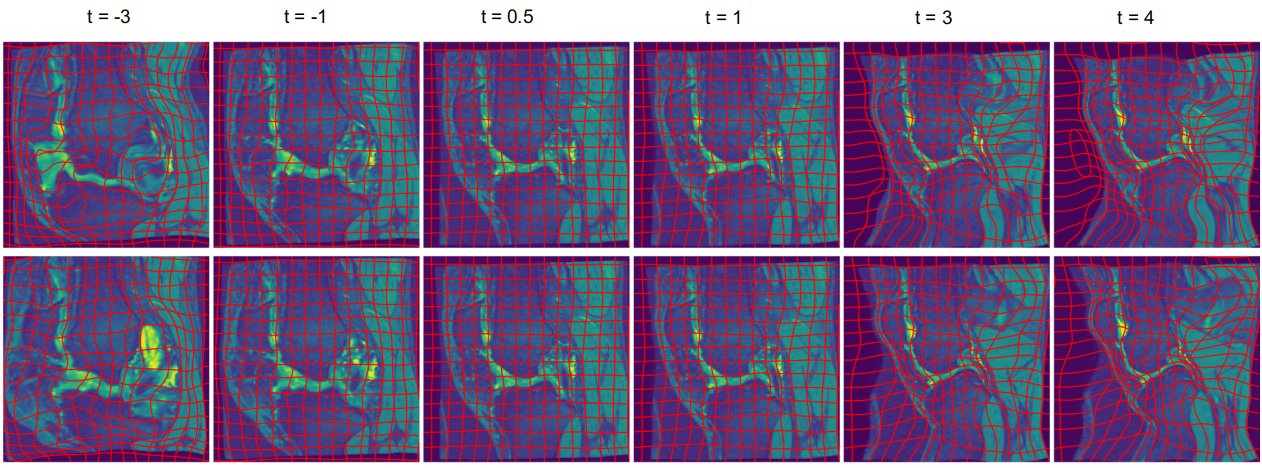}
\caption{Comparison between inter- and extrapolation of the displacement field (top row) and geodesic inter- and extrapolation (bottom row). We show the center slices from the sagittal view of the 3D MRI knee images. For both methods, we assume they have the same transformation $\varphi^{-1}(1)$ at $t=1$. Then we compute the displacement-field based inter- and extrapolation as $\varphi^{-1}_{affine}(t,x)=(\varphi^{-1}(1,x)-x) t + x$, whereas $\varphi^{-1}_{LDDMM}$ is obtained via geodesic shooting (based on solving the EPDiff Eq.~\ref{eq:lddmm_epdiff}). For large deformations, i.e., $t=-3$ and $t=4$, affine extrapolation results in foldings while extrapolation via the LDDMM geodesic results in diffeomorphic transformations.}
\label{fig:fluid_vs_linear}
\end{figure}


\begin{figure}[!htb]
\vspace{-1.2cm}
\includegraphics[width=1\textwidth]{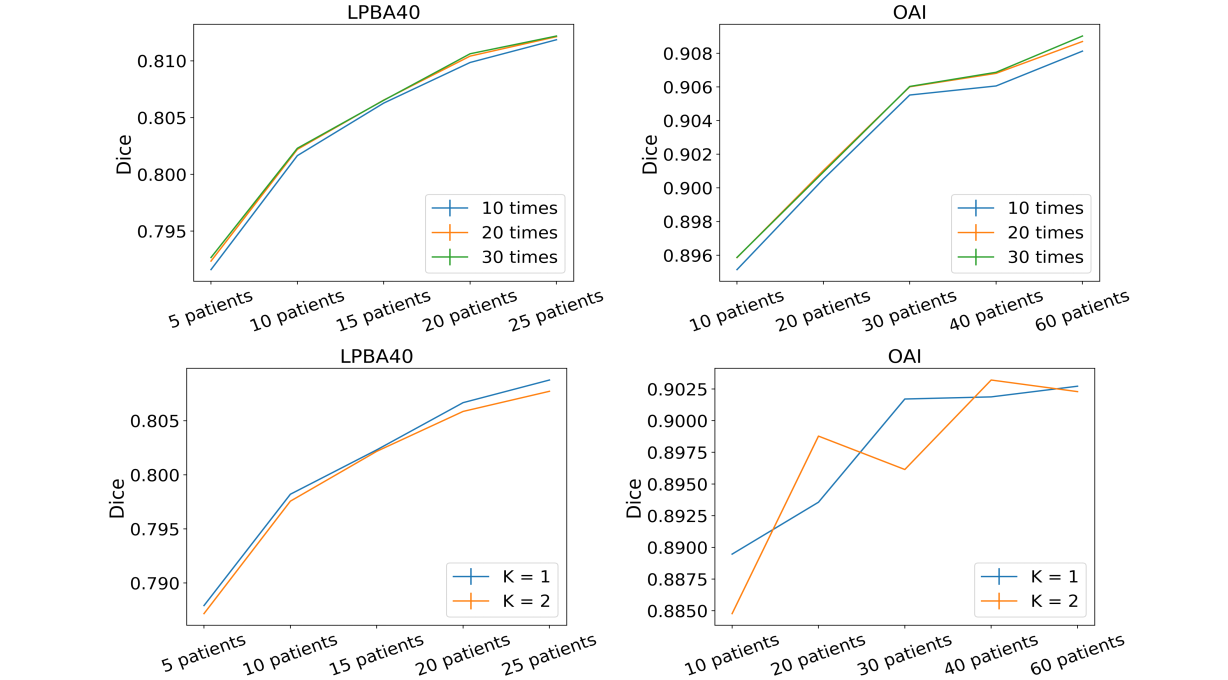}
\caption{Ablation study on testing augmentation size (first row) and the choice of K (second row). For the first row, we evaluate segmentation accuracies for a different numbers of augmentation samples in the test phase. $N$ times denotes that a testing image is deformed by $N$ different random transformations drawn from the geodesic subspace. Segmentation accuracies start to saturate for $N\geq 20$. For the second row, we evaluate the performance for different choices of $K$ (i.e., the dimensionality of the geodesic subspace). A larger $K$ does not necessarily result in better segmentation accuracies.
}
\label{fig:post_ens}
\end{figure}

\enlargethispage{3\baselineskip}

\begin{figure}[!t]
\centering
\vspace{-0.3cm}
\includegraphics[width=0.95\textwidth]{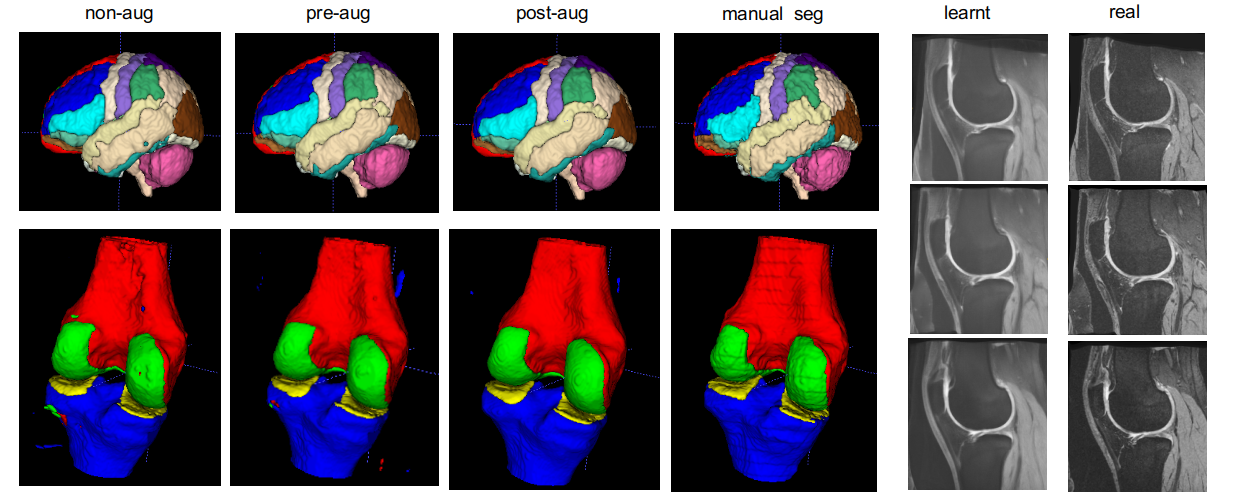}
\vspace{-0.4cm}
\caption{ From the first to the fourth column, we visualize the segmentation results in Sec.~\ref{sec:segmentation} on LPBA40 (first row) and OAI (second row); from left to right: results without augmentation (non-aug), training phase augmentation (pre-aug), testing phase augmentation (post-aug), and the manual segmentations. We observe segmentation refinement after {\it pre-aug} and {\it post-aug}. For the last two columns, we compare the learnt appearance $A+r_i$ (fifth column) and the real appearance $I_i \circ \varphi_i$ (sixth column) in Sec.~\ref{sec:one_shot_segmentation}, where each row is a patient. The learnt appearance is smoother and hence shows less image texture. In our case, the segmentation network trained using the learnt appearance does not match the noisy testing data well.}
\label{fig:vis}
\end{figure}


\begin{figure}[!h]
\vspace{-0.1cm}
\centering
\includegraphics[width=0.9\textwidth, height=0.7\textwidth]{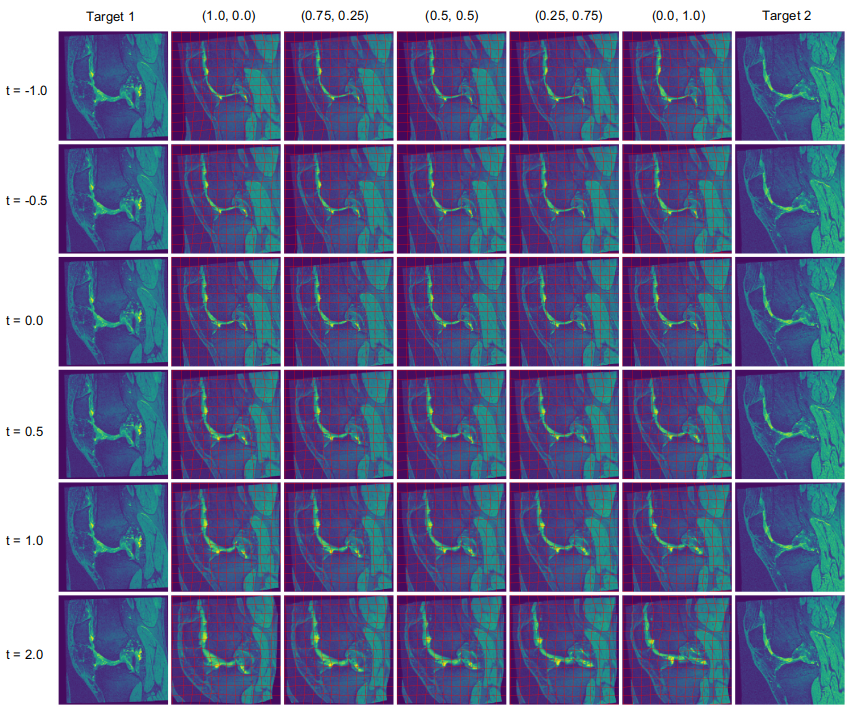}
\vspace{-0.3cm}
\caption{Visualization of synthesized images obtained from a source image and its $K=2$ geodesic subspace. From left to  right: the first and the last column show the two target images. The remaining columns show synthesized images $I_c \circ \varphi^{-1}_{\tilde m_0^c}(t)$. The column index $(\cdot,\cdot)$ denotes the mixture weights $(\lambda_1, \lambda_2)$ in Eq.~\ref{eq:convex}, determining how much target 1 and target 2 influence the overall deformation. From top to bottom: each row refers to the time $t$ sampled along the geodesic path. For the row $t=0$, the transformation is the identity, i.e., $I_c= I_c \circ \varphi^{-1}_{\tilde m_0^c}(0)$. For the row $t=1$, the warped image has similar anatomical structure as target 1 when $(\lambda_1, \lambda_2)=(1.0,0.0)$ while is similar to target 2 when  $(\lambda_1, \lambda_2)=(0.0,1.0)$. Columns $(1.0,0.0)$ and $(0.0,1.0)$ show samples on two geodesic paths ($K=1$) toward target 1 and target 2, respectively.
}
\vspace{-2cm}
\label{fig:2d_case1}
\end{figure}

\clearpage
%
%

\end{document}